\def\year{2022}\relax
\documentclass[letterpaper]{article} % DO NOT CHANGE THIS
\usepackage{aaai22}  % DO NOT CHANGE THIS
\usepackage{times}  % DO NOT CHANGE THIS
\usepackage{helvet}  % DO NOT CHANGE THIS
\usepackage{courier}  % DO NOT CHANGE THIS
\usepackage[hyphens]{url}  % DO NOT CHANGE THIS
\usepackage{graphicx} % DO NOT CHANGE THIS
\usepackage{natbib}  % DO NOT CHANGE THIS AND DO NOT ADD ANY OPTIONS TO IT
\usepackage{caption} % DO NOT CHANGE THIS AND DO NOT ADD ANY OPTIONS TO IT
\DeclareCaptionStyle{ruled}{labelfont=normalfont,labelsep=colon,strut=off} % DO NOT CHANGE THIS
\usepackage{algorithm}
\usepackage{algorithmic}

\usepackage{newfloat}
\usepackage{listings}
\usepackage{multirow}
\usepackage{booktabs}
\usepackage{amsmath}
\usepackage{amssymb}
\usepackage{bbding}
\usepackage{bm}
\floatstyle{ruled}
\newfloat{listing}{tb}{lst}{}
\floatname{listing}{Listing}

\title{Self-Supervised Video Object Segmentation by Motion-Aware Mask Propagation}
\author {
    Bo Miao\textsuperscript{\rm 1},
    Mohammed Bennamoun\textsuperscript{\rm 1},
    Yongsheng Gao\textsuperscript{\rm 2},
    Ajmal Mian \textsuperscript{\rm 1}
}
\affiliations {
    \textsuperscript{\rm 1} The University of Western Australia\\
    \textsuperscript{\rm 2} Griffith University\\
    bo.miao@research.uwa.edu.au, \{mohammed.bennamoun, ajmal.mian\}@uwa.edu.au, yongsheng.gao@griffith.edu.au
}

\begin{document}

\maketitle

% TODO: Format, e.g. ytb 2018 2019
\begin{abstract}

We propose a self-supervised spatio-temporal matching method, coined Motion-Aware Mask Propagation (MAMP), for video object segmentation. MAMP leverages the frame reconstruction task for training without the need for annotations. During inference, MAMP extracts high-resolution features from each frame to build a memory bank from the features as well as the predicted masks of selected past frames. MAMP then propagates the masks from the memory bank to subsequent frames according to our proposed motion-aware spatio-temporal matching module to handle fast motion and long-term matching scenarios. Evaluation on DAVIS-2017 and YouTube-VOS datasets show that MAMP achieves state-of-the-art performance with stronger generalization ability compared to existing self-supervised methods, i.e., 4.2\% higher mean $\mathcal{J}\&\mathcal{F}$ on DAVIS-2017 and 4.85\% higher mean $\mathcal{J}\&\mathcal{F}$ on the unseen categories of YouTube-VOS than the nearest competitor. Moreover, MAMP performs at par with many {\em supervised} video object segmentation methods.
\end{abstract}

\section{Introduction}

Video object segmentation (VOS) is a fundamental problem in visual understanding where the aim is to segment objects of interest from the background in unconstrained videos. VOS enables machines to sense the motion pattern, location, and boundaries of the objects of interest in videos, which is useful in a wide range of applications. For example, in video editing, manual frame-wise segmentation is laborious and does not maintain temporal consistency whereas VOS can segment all frames automatically using the mask of one frame as a guide. The problem of segmenting objects of interest in a video using the ground truth object masks provided for only the first frame is referred to as semi-supervised VOS. This is challenging because the appearance of objects in a video change significantly due to fast motion, occlusion, scale variation, etc. Moreover, other similar looking non-target objects may confuse the model to segment incorrect objects.

\begin{figure}[t!]
\centering
\includegraphics[width=\columnwidth]{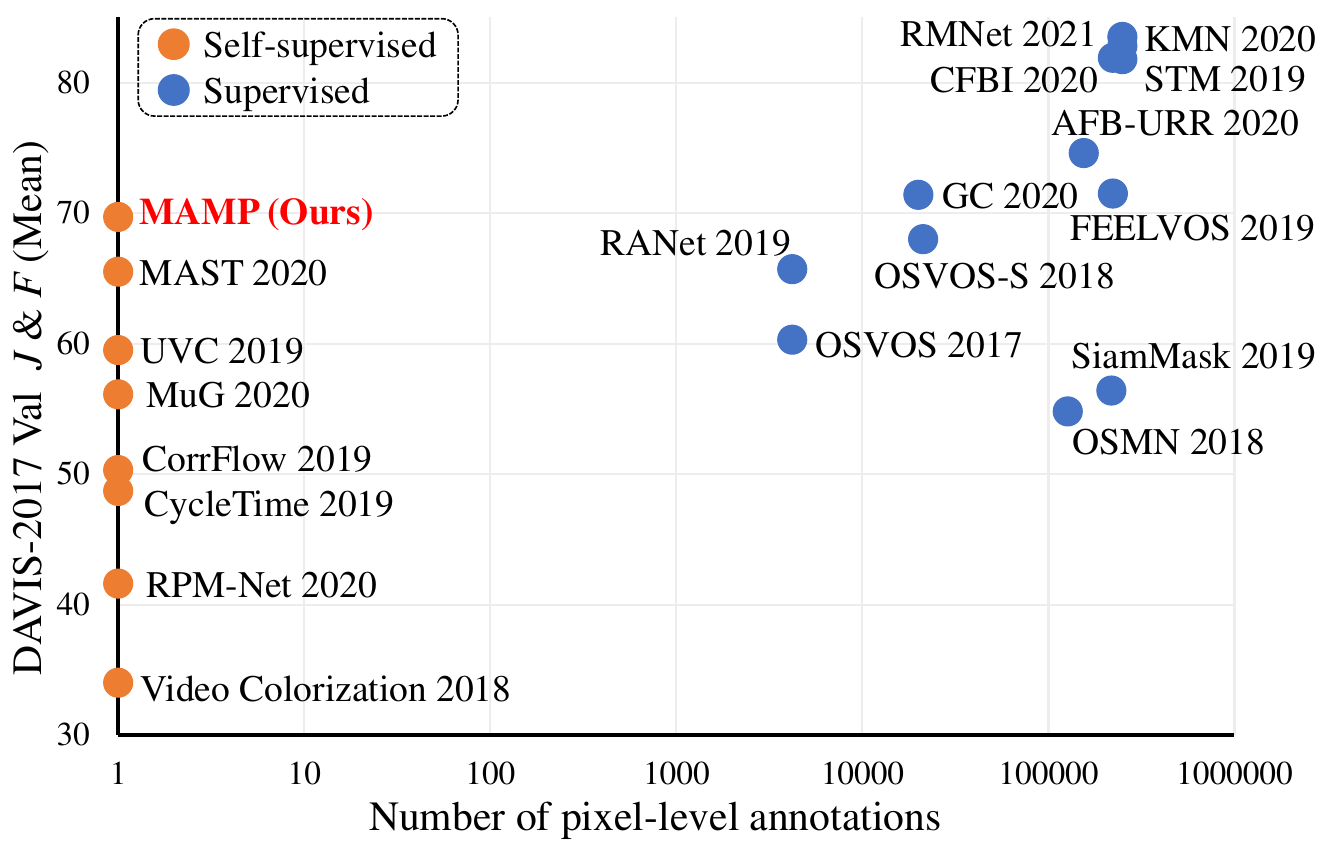}
\caption{Comparison on DAVIS-2017 validation set with other methods. MAMP outperforms existing self-supervised methods, and is at par with some supervised methods trained with large amounts of annotated data.}
\label{fig:1}
\end{figure}

Semi-supervised VOS techniques fall into two categories: supervised and self-supervised. Supervised approaches \cite{STM,CFBI} use rich annotation information from training data to learn the model achieving great success in VOS. However, these methods are unattractive given their reliance on accurate pixel-level annotations for training (see Fig.~\ref{fig:1}), which are expensive to generate. Moreover, supervised approaches  struggle to maintain the same performance in the wild. In contrast, self-supervised methods \cite{MuG,MAST} learn feature representations based on the intrinsic properties of the video frames, and thus do not require any annotations and can better generalize to unseen objects. Even though the motivations behind existing self-supervised methods are different, they share the same objective of learning to extract general feature representations and construct precise spatio-temporal correspondences to propagate the object masks in video sequences. Exploiting the spatio-temporal coherence in videos, the pretext tasks of self-supervised methods can be designed to either maximize the temporal cycle-correspondence consistency \cite{MuG} or minimize the reconstruction or prediction loss \cite{MAST}. Once trained, the models are able to extract general feature representations and build spatio-temporal correspondences between the reference and query frames. Therefore, pixels in the query frames can be classified according to the mask labels of their corresponding region of interests (ROI) in the reference frames. Despite their simplicity, existing self-supervised methods perform poorly in cases of fast motion and long-term matching scenarios.

To overcome the above challenges, we propose a self-supervised method coined motion-aware mask propagation (MAMP). Similar to previous self-supervised methods, MAMP learns image features and builds spatio-temporal correspondences without any annotations during training. During inference, MAMP first leverages the feature representations and the given object masks for the first frame to build a memory bank. The proposed motion-aware spatio-temporal matching module in MAMP then exploits the motion cues to mitigate the issues caused by fast motion and long-term correspondence mismatches, and propagates the mask from the memory bank to subsequent frames. Moreover, the proposed size-aware image feature alignment module fixes the misalignment during mask propagation and the memory bank is constantly updated by the past frames to provide the most appropriate spatio-temporal guidance. We evaluate MAMP on the DAVIS-2017 and YouTube-VOS benchmarks to verify its effectiveness as well as generalization ability.
Our contributions are summarized as follows:
\begin{itemize}
\item[$\bullet$] We propose Motion-Aware Mask Propagation (MAMP) for VOS that trains the model end-to-end without any annotations and effectively propagates the masks across frames.
\item[$\bullet$] We propose a motion-aware spatio-temporal matching module to mitigate errors caused by fast motion and long-term correspondence mismatches. This module improves the performance of MAMP on YouTube-VOS by 6.4\%.
\item[$\bullet$] Without any bells and whistles (e.g., fancy data augmentations, online adaptation, and external datasets), MAMP significantly outperforms existing self-supervised methods by 4.2\% on DAVIS-2017 and 4.85\% on the unseen categories of YouTube-VOS.
\item[$\bullet$] Experiment on YouTube-VOS dataset shows that MAMP has the best generalization ability compared to existing self-supervised and supervised methods.
\end{itemize}

\begin{figure*}[t!] %  figure placement: here, top, bottom, or page
\centering
\includegraphics[width=0.9\textwidth]{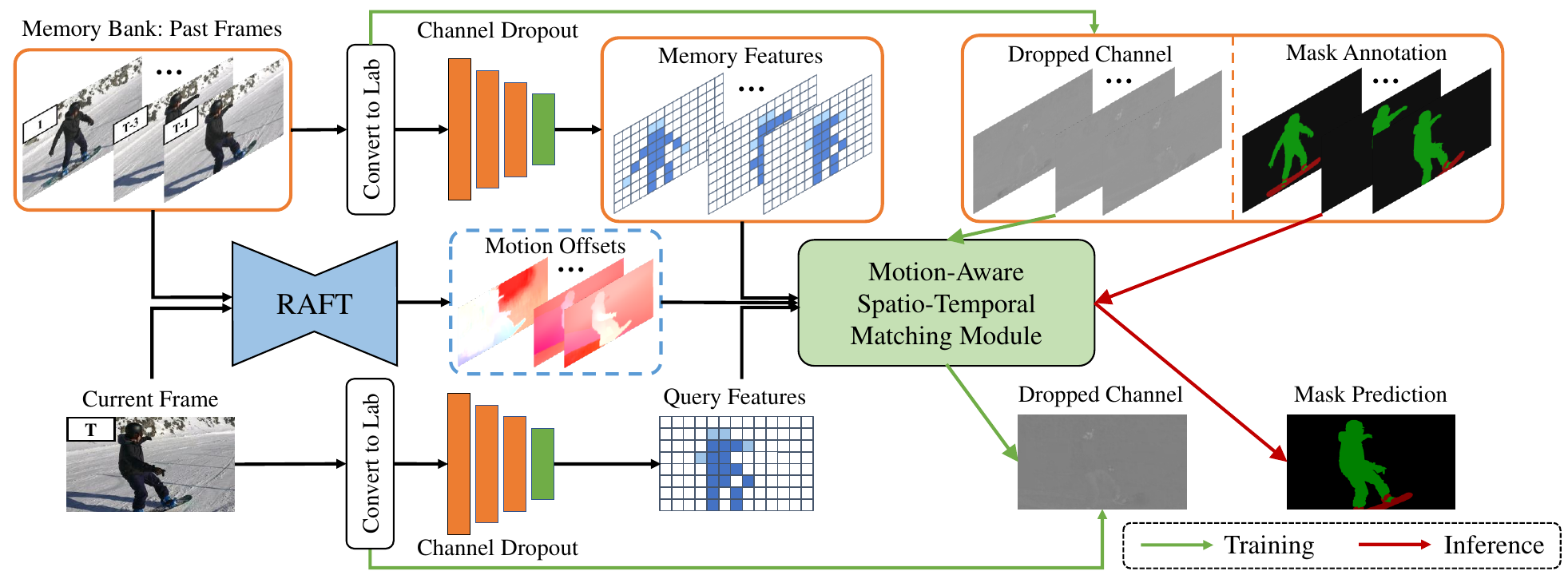}
\caption{{Framework of the proposed MAMP. During training, a random pair of neighboring video frames is sampled. The frames are converted to $Lab$ color space and channel dropout is used only on the $ab$ channels to generate the reconstruction target for self-supervision. During inference, a parameter sharing encoder is used to encode the frames into feature maps. The proposed motion-aware spatio-temporal matching module is used to compute spatio-temporal correspondences between the past frames in the memory bank and the current frame and propagate the masks across frames.}}
\label{fig:2}
\end{figure*}

\section{Related Work}

{\noindent\textbf{Semi-supervised VOS} aims to leverage the ground truth object mask given (only) in the first frame to segment the objects of interest in subsequent frames. Existing semi-supervised VOS methods can be divided into online-learning and offline-learning methods depending on whether online adaptation is needed during inference. The former usually update the networks dynamically during inference based on the first frame \cite{OSVOS,OSVOS_S,MaskTrack,TAN_DTTM}, synthetic frames \cite{Lucid,PReMVOS}, or multiple selected past frames \cite{OnAVOS,e_osvos} of each video making the networks object-specific. Literature has shown the effectiveness of online-learning \cite{Ranet}, however, it is time consuming and adversely affects the models' generalization ability. Offline-learning methods usually propagate the given object mask or features of the first frame either explicitly or implicitly to subsequent frames, making the expensive online-adaptation no longer necessary \cite{OSMN,SiamMask}. Recurrent neural networks \cite{RVOS,DyeNet} and recurrent training strategies \cite{RGMP} can implicitly propagate the spatio-temporal features from past frames to the current frame. Memory networks with attention mechanism \cite{STM,AFB_URR,GC,KMN,rmnet}, correlation mechanism \cite{Videomatch,FEELVOS,CFBI,MAST}, or clustering methods \cite{PML} can explicitly propagate the spatio-temporal features from multiple different past frames to the current frame. Existing offline-learning methods usually perform local or non-local spatio-temporal matching for temporal association and mask propagation. However, non-local matching is noisy and has a large memory footprint, while local matching struggles to cope with problems from fast motion and long-term correspondence mismatches.}

The proposed MAMP is an offline-learning method and does not require time-consuming online adaptation. MAMP leverages a dynamically updated memory bank to store features and masks from selected past frames, and propagates masks effectively according to our proposed motion-aware spatio-temporal matching module. Unlike previous matching methods, our motion-aware spatio-temporal matching module not only excludes the noisy matching results but also mitigates the problems caused by fast motion and long-term correspondence mismatches.

{\noindent\textbf{Memory Networks}} aim to capture the long-term dependencies by storing temporal features or different categories of features in a memory module. LSTM \cite{LSTM} and GRU \cite{GRU} implicitly represent spatio-temporal features with local memory cells in a highly compressed way limiting the representation ability. Memory networks \cite{MemNet} were introduced to explicitly store the important features. A common memory network in VOS is STM \cite{STM} which incrementally adds the features of past frames to the memory bank, and leverages the non-local spatio-temporal matching to provide spatio-temporal features. However, the incremental memory bank updates are impractical when segmenting long videos due to the growing memory cost. In this work, we divide the memory into long-term and short-term memory. The former is fixed, whereas the latter is updated dynamically using the past few frames making our MAMP memory efficient.

\noindent\textbf{Self-supervised Learning}
can learn general feature representations and spatio-temporal correspondences based on the intrinsic properties of videos. It has shown promising capacity on various downstream tasks as it does not require annotations and can better generalize \cite{Video_Color,PreDecode,UVC,CubicPuzz,FacePred,InterIntra,VideoMoCo}. Many pretext tasks have been explored for self-supervised learning such as future frame prediction \cite{FuFramePred}, query frame reconstruction \cite{CorrFlow,RPMNet,MAST}, patch re-localization \cite{CycleTime,MuG}, and motion statistics prediction \cite{MotionStat}.

\section{Method}
\subsection{Overview}
Fig.~\ref{fig:2} shows an overview of our method for VOS. MAMP is trained with the reconstruction task to learn feature representations and construct robust spatio-temporal correspondences between pairs of frames from the same video. Hence, \emph{zero} annotation is required to train the model. During inference, MAMP segments the frames in a sequential manner. The parameter sharing encoder is used to extract frame-wise features and the motion-aware spatio-temporal matching module propagates the masks from the past frames to the current frame according to the spatio-temporal affinity matrix. The size-aware image feature alignment module also facilitates the mask propagation to prevent misalignment.

\subsection{Self-supervised Feature Representation Learning}
We use the reconstruction task for self-supervised feature representation learning and robust spatio-temporal matching. Since the channel correlation in $Lab$ color space is smaller than that of $RGB$ \cite{ColorTransfer}, we randomly dropout one of the $ab$ channels and make the $Lab$ color as the reconstruction target. Dropout preserves enough information for the input while avoiding trivial solutions. Therefore, the model learns the general feature representations and the spatio-temporal correspondences between the reference and query frames instead of learning how to predict the missing channel from the observed channels. To minimize the reconstruction loss, semantically similar pixels between reference and query frames are forced to have highly correlated feature representations, while semantically dissimilar pixels are forced to have weakly correlated feature representations. Finally, the reconstruction target of the query frames is predicted according to the highly correlated ROIs in the reference frames.

Specifically, given a reference and query frame $\{\bm{I}_{r},\bm{I}_{q}\} \in{\mathbb{R}^{H\times W\times 3}}$ from a video, a parameter-sharing convolutional encoder $\Phi(g(\bm{I}); \theta)$ is used to extract their feature representations $\{\bm{F}_{r},\bm{F}_{q}\}\in{\mathbb{R}^{h\times w\times c}}$, where $g (\cdot)$ is the information bottleneck to prevent trivial solutions. The dropped channels of the two frames $\{\bm{C}_{r}, \bm{C}_{q}\} \in{\mathbb{R}^{H\times W\times 1}}$ are downsampled to the resolution of the features $\{\bm{C}_{r,d}, \bm{C}_{q,d}\} \in{\mathbb{R}^{h\times w\times 1}}$ based on the size-aware image feature alignment module.

To enable $\bm{C}_{r}$ to represent and reconstruct $\bm{C}_{q}$, the spatio-temporal affinity matrix $\bm{A}_{q,r}\in {\mathbb{R}^{hw\times R}}$ that represents the strength of the correlation between $\bm{F}_{q}$ and $\bm{F}_{r}$ is:
$$\bm{A}_{q,r}^{i,j} = \frac{exp(\langle \bm{F}_{q}^{i},\bm{F}_{r}^{j}\rangle/\sqrt{c})}{\sum_{j\in{\bm{R}}} exp(\langle \bm{F}_{q}^{i},\bm{F}_{r}^{j}\rangle/\sqrt{c})}, \eqno{(1)}$$
where $i$ and $j$ are the locations in $\bm{F}_{q}$ and $\bm{F}_{r}$, respectively. $\langle \cdot , \cdot \rangle$ is the dot product between two vectors, and $c$ refers to the number of channels to re-scale the correlation value and $\bm{R}$ is the ROI of $i$. Next, a location $i$ in $\bm{C}_{q,d}$ is represented by the weighted sum of the corresponding ROI in $\bm{C}_{r,d}$:

$$\bm{\widehat{C}}_{q,d}^{i} = \sum_{j\in{\bm{R}}} \bm{A}_{q,r}^{i,j} \bm{C}_{r,d}^{j}  \eqno{(2)}$$

Finally, $\bm{\widehat{C}}_{q,d}^{i}$ is upsampled to $\bm{\widehat{C}}_{q}^{i}$, and Huber Loss $\mathcal{L}$ is used to force $\bm{\widehat{C}}_{q}^{i}$ to be close to $\bm{C}_{q}^{i}$:
$$\mathcal{L} = \frac{1}{n} \sum_{i=1}^{N} \bm{\mathcal{Z}}_{i} \eqno{(3)}$$
\noindent where
$$\bm{\mathcal{Z}}_{i} = \begin{cases}
            0.5(\bm{\widehat{C}}_{q}^{i}-\bm{C}_{q}^{i})^{2},\ \ \ {\rm if} \ \vert \bm{\widehat{C}}_{q}^{i}-\bm{C}_{q}^{i} \vert \textless 1 \\
            \vert \bm{\widehat{C}}_{q}^{i}-\bm{C}_{q}^{i}\vert-0.5 ,\ {\rm otherwise}.
            \end{cases} \eqno{(4)}$$

\subsection{Motion-aware Spatio-temporal Matching}

\begin{figure}[t!] %  figure placement: here, top, bottom, or page
\centering
\includegraphics[width=\columnwidth]{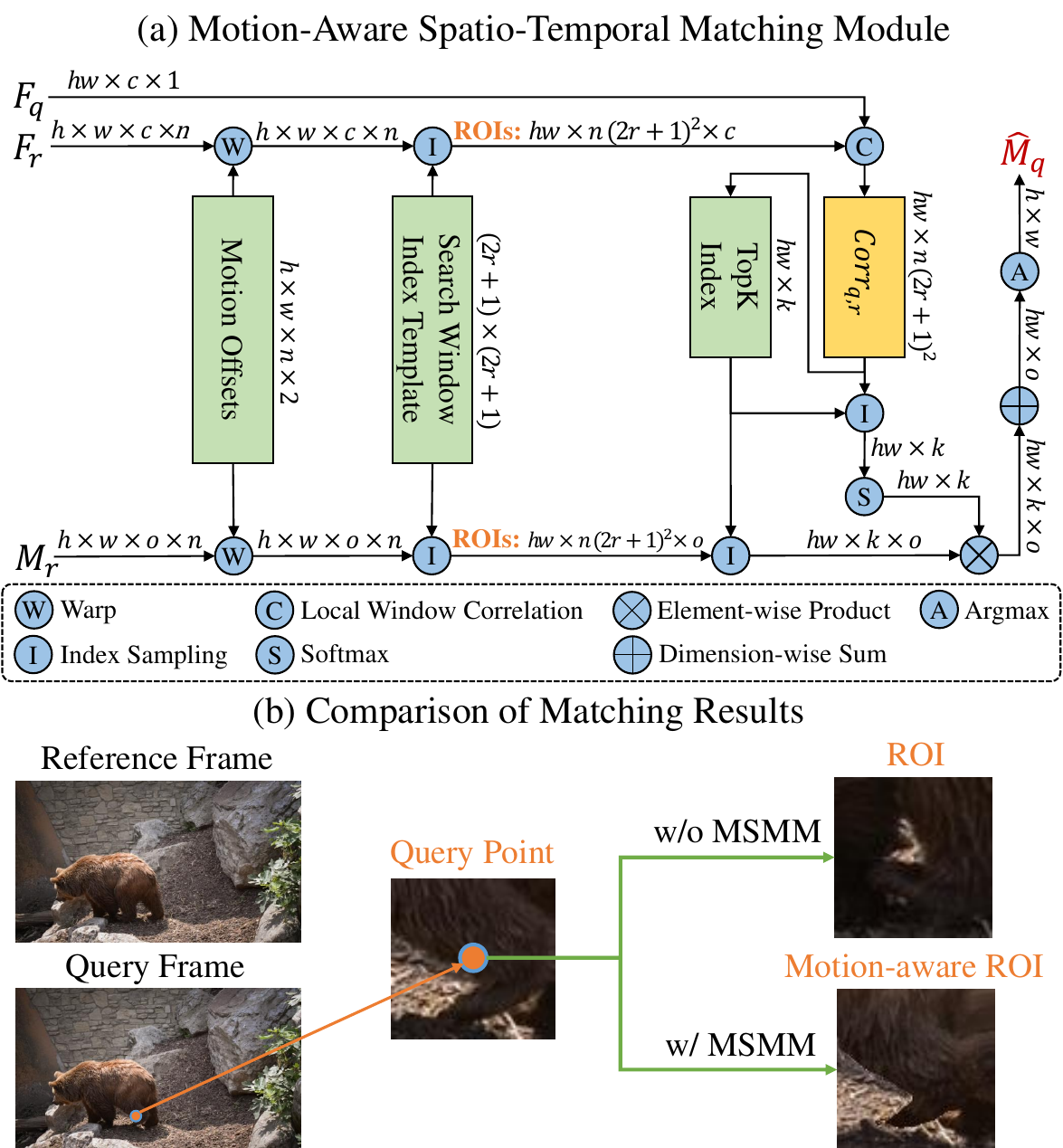}
\caption{(a) Proposed motion-aware spatio-temporal matching module. $\bm{F}_{q}$, $\bm{F}_{r}$ are the features of query, reference frames. $\bm{M}_{r}$ is the masks of reference frames, and $\bm{\widehat{M}}_{q}$ is the predicted mask. (b) Comparison of the matching results between the vanilla local spatio-temporal matching module and our motion-aware spatio-temporal matching module.}
\label{fig:3}
\end{figure}

To account for moving objects, VOS models should be able to retrieve the corresponding ROIs from the reference frames for mask propagation under motion. To meet this constraint, non-local spatio-temporal matching methods \cite{STM} consider all locations in the reference frames as potential ROIs. However, non-local matching generates many noisy matches and has a large memory footprint. Local spatio-temporal matching methods \cite{MAST} retrieve the ROIs in the reference frames based on the location coordinates in the query frame and a pre-defined retrieval radius. Although local matching is more efficient, existing local methods have limited receptive fields limiting their ability to localize the most correlated ROIs when they encounter fast motion or after long-term matching.

We propose a motion-aware spatio-temporal matching module that leverages optical flow to enable the query locations to retrieve the most correlated ROIs from the reference frames, see Fig.~\ref{fig:3}(a). We use RAFT \cite{RAFT} which costs only about 18ms and 13ms to compute the motion offsets between frame pairs in DAVIS-2017 and YouTube-VOS datasets, respectively. As shown in Fig.~\ref{fig:3}(b), the vanilla local spatio-temporal matching method cannot retrieve the most correlated ROIs for the query point. However, with our motion-aware spatio-temporal matching, the query point can find its most correlated ROIs even if the ROI pixels are not consecutive in raw space.

Our motion-aware spatio-temporal matching module takes the features of a query frame $\bm{F}_{q} \in{\mathbb{R}^{h\times w\times c}}$, the features of reference frames $\bm{F}_{r} \in{\mathbb{R}^{h\times w\times c\times n}}$, the downsampled masks of reference frames $\bm{M}_{r} \in{\mathbb{R}^{h\times w\times o\times n}}$, and the downsampled motion offsets $\bm{MO}(\Delta x, \Delta y) \in \mathbb{R}^{h \times w \times n \times 2}$ between the query and reference frames as inputs, where $\Delta x$ and $\Delta y$ are the displacement vectors along the horizontal and vertical directions, respectively. $\bm{F}_{r}$ and $\bm{M}_{r}$ are first warped according to $\bm{MO}(\Delta x, \Delta y)$ making the locations with the same coordinates in $\bm{F}_{q}$ and the warped $\bm{F}_{r}$ and $\bm{M}_{r}$ to be the most similar pairs. Therefore, for one location $i \in \bm{F}_{q}(x,y)$ in the query frame, the initial corresponding ROIs can be retrieved, i.e., $\bm{R}=\{\forall j\in{\bm{F}_{r}}, \vert w(j)-i\vert \leq r\}$, where $r$ is the radius of ROIs and $w(j) \in \bm{F}_{r}(x+\Delta x,y+\Delta y)$. Due to different input resolutions, we empirically set $r$ to 6 during training and 12 during inference. Subsequently, the initial spatio-temporal affinity matrix $\bm{Corr}_{q,r} \in{\mathbb{R}^{h\times w\times n(2r+1)^{2}}}$is computed and the TopK selection block is used to filter out the weakly correlated locations in the ROIs to save the Softmax operation from being affected by noise. $k$ is set to 36 for all experiments (see Supplementary Material). Finally, mask propagation is achieved by multiplying the selected spatio-temporal affinity matrix with the corresponding ROIs in $\bm{M}_{r}$.

During inference, with all the ROIs being dynamically sampled from different reference frames in the memory bank based on the motion-aware spatio-temporal matching module, the mask propagation becomes more accurate and the problems caused by fast motion and long-term correspondence mismatches are alleviated.

\subsection{Size-aware Image Feature Alignment}
\begin{figure}[t] %  figure placement: here, top, bottom, or page
\centering
\includegraphics[width=\columnwidth]{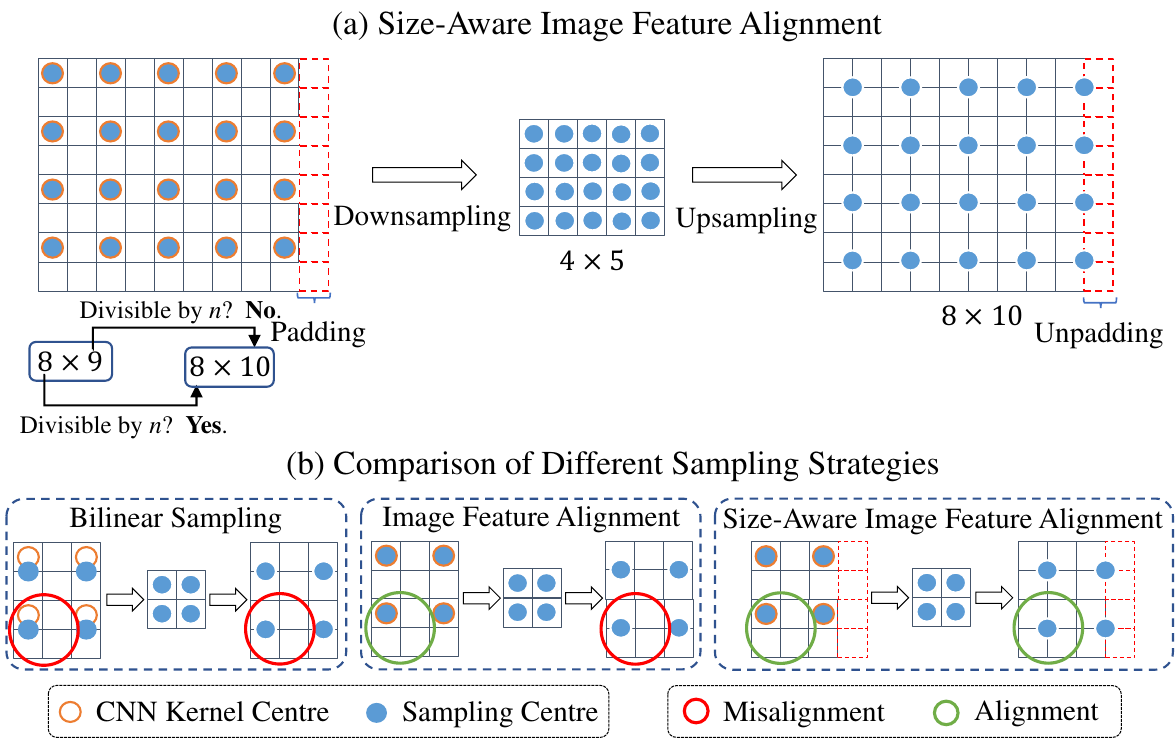}
\caption{Size-aware image feature alignment module in comparison to bilinear sampling and image feature alignment. \emph{n} is the ratio of the input to downsampled size. The proposed size-aware image feature alignment fixes the misalignment in both downsampling and upsampling stages.} % up-sample center point should on the centre because it is the convolutional centre
\label{fig:4}
\end{figure}

To reduce memory consumption, previous methods perform bilinear downsampling on the supervision signals (masks) of the reference frames and propagate these signals at the feature resolution. However, this operation introduces misalignment (see Fig.~\ref{fig:4}(b)) between the strided convolution layers and the supervision signals from na\"ive bilinear downsampling and upsampling. MAST \cite{MAST} has an image feature alignment module to deal with this problem but it does not cater for the misalignment caused at the upsampling stage. To solve this problem, we propose a size-aware image feature alignment module where the supervision signals are automatically padded (see Fig.~\ref{fig:4}(a)) so that the input size can be divisible by the size after downsampling, and the supervision signals are sampled at the convolution kernel centers. Hence, the misalignment is removed at both downsampling and upsampling stages.

\begin{table*}[t!]
\small
\centering
\begin{tabular}{lllrlrcccc}
\toprule[1.5pt]
Method & Year & Backbone & Param. & T. Data & Vid. Length & Supervised & \( \mathcal{J} \)\&\( \mathcal{F} \) (Mean) & \( \mathcal{J} \) (Mean) & \( \mathcal{F} \) (Mean)\\
\hline
Vid. Color. & 2018 & ResNet-18 & 5M & K & 800 hrs & \XSolidBrush & 34.0 & 34.6 & 32.7 \\
CycleTime & 2019 & ResNet-50 & 9M & V & 344 hrs & \XSolidBrush & 48.7 & 46.4 & 50.0 \\
CorrFlow & 2019 & ResNet-18 & 5M & O & 14 hrs & \XSolidBrush & 50.3 & 48.4 & 52.2 \\
UVC & 2019 & ResNet-18 & 3M & K & 800 hrs & \XSolidBrush & 59.5 & 57.7 & 61.3 \\
RPM-Net & 2020 & ResNet-101 & 43M & DY & 5.67 hrs & \XSolidBrush & 41.6 & 41.0 & 42.2 \\
Mug & 2020 & ResNet-50 & 9M & O & 14 hrs & \XSolidBrush & 56.1 & 54.0 & 58.2 \\
MAST & 2020 & ResNet-18 & 5M & Y & 5.58 hrs & \XSolidBrush & 65.5 & 63.3 & 67.6 \\
\textbf{Ours} & 2021 & ResNet-18 & 5M & Y & 5.58 hrs & \XSolidBrush & \textbf{69.7} & \textbf{68.3} & \textbf{71.2} \\
\hline
OSVOS & 2017 & VGG-16 & 15M & ID & & \CheckmarkBold & 60.3 & 56.6 & 63.9 \\
OSMN & 2018 & VGG-16 & 15M & ICD & & \CheckmarkBold & 54.8 & 52.5 & 57.1 \\
OSVOS-S & 2018 & VGG-16 & 15M & IPD & & \CheckmarkBold & 68.0 & 64.7 & 71.3 \\
%PReMVOS & 2018 & ResNet-101 & 43M & ICPMD & & \CheckmarkBold & 77.8 & 73.9 & 81.8 \\
SiamMask & 2019 & ResNet-50 & 9M & ICY & & \CheckmarkBold & 56.4 & 54.3 & 58.5 \\
FEELVOS & 2019 & Xception-65 & 38M & ICDY & & \CheckmarkBold & 71.5 & 69.1 & 74.0 \\
STM & 2019 & ResNet-50 & 9M & ICPSEDY & & \CheckmarkBold & 81.8 & 79.2 & 84.3 \\
GC & 2020 & ResNet-50 & 9M & ISEHD & & \CheckmarkBold & 71.4 & 69.3 & 73.5 \\
AFB-URR & 2020 & ResNet-50 & 9M & ICPSED & & \CheckmarkBold & 74.6 & 73.0 & 76.1 \\
%KMN & 2020 & ResNet-50 & 9M & ICPSEDY & & \CheckmarkBold & 82.8 & 80.0 & 85.6 \\
CFBI & 2020 & ResNet-101 & 43M & ICDY & & \CheckmarkBold & 81.9 & 79.1 & 84.6 \\
RMNet & 2021 & ResNet-50 & 9M & ICPSEDY & & \CheckmarkBold & \textbf{83.5} & \textbf{81.0} & \textbf{86.0} \\
\toprule[1.5pt]
\end{tabular}
\caption{Evaluation on DAVIS-2017 validation set. Note that each method modifies vanilla backbone models to suit their framework. Training Dataset (T. Data) notations: C=COCO, D=DAVIS, E=ECSSD, H=HKU-IS, I=ImageNet, K=Kinetics, M=Mapillary, O=OxUvA, P=PASCAL-VOC, S=MSRA10K, V=VLOG, Y=YouTube-VOS.}
\label{tabdaviseval}
\end{table*}

\section{Implementation Details}

\textbf{Training:} We modify ResNet-18 and use it as the encoder to extract image features with a spatial resolution of 1\verb|/|4 of the input images (see Supplementary Material for details). The encoder parameters are randomly initialized \emph{without} pre-training. A pair of nearby video frames are randomly sampled as the reference and query frame, and the reconstruction task with Huber Loss is used to train the model. For pre-processing, the frames are resized to 256 $\times$ 256, and \emph{no} data augmentation is used. We train our model with pairwise frames for 33 epochs on YouTube-VOS using a batch-size of 24 for all experiments. We adopt Adam optimizer with the base learning rate of 1$e$-3, and the learning rate is divided by 2 after 0.4M, 0.6M, 0.8M, and 1.0M iterations, respectively. Our model is trained end-to-end \emph{without} any multi-stage training strategies. The training takes about 11 hours on one NVIDIA GeForce 3090 GPU.

\noindent\textbf{Testing:} The proposed MAMP does \emph{not} require time-consuming online adaption to fine-tune the model during testing. To be consistent with benchmarks, MAMP is evaluated on the YouTube-VOS validation set at half resolution and DAVIS-2017 validation set at full resolution. Results on DAVIS-2017 and YouTube-VOS are obtained using the official evaluation code and server. During testing, MAMP leverages the size-aware image feature alignment module to fix the misalignment, and uses the trained encoder to extract features from each frame. After that, MAMP uses the proposed motion-aware spatio-temporal matching to propagate the masks from the memory bank to subsequent frames. The memory bank of MAMP is updated dynamically to include $I_{0}$ and $I_{5}$ as long-term memory and $I_{t-5}$, $I_{t-3}$, and $I_{t-1}$ as short-term memory.

\section{Experiments}
\subsection{Datasets}
\noindent\textbf{DAVIS-2017} \cite{davis2017} is commonly for VOS in \emph{short} video clips and complex scenes. It contains 150 videos with over 200 objects. The validation set of DAVIS-2017 contains 30 videos.

\noindent\textbf{YouTube-VOS} \cite{s2s} is the largest dataset for VOS with \emph{long} video clips. It contains over 4000 high-resolution videos with 7000+ objects. The validation set of YouTube-VOS 2018 contains 474 videos. Unlike previous methods \cite{STM,rmnet} that leverage several external datasets to train the model, we only train our model on YouTube-VOS and test our model on both DAVIS-2017 and YouTube-VOS. Unless specified otherwise, the YouTube-VOS dataset in this paper refers to the 2018 version which is consistent with previous benchmarks.

\begin{table*}[h]
\small
\centering
\begin{tabular}{l|cccc|cccccccc}
\toprule[1.5pt]
Method & Vid. Color. & CorrFlow & MAST & \textbf{Ours} & OSMN & RGMP & OnAVOS & S2S & A-GAME & STM & GC & RMNet \\
% \hline
% Year & 2018 & 2019 & 2020 & 2021 & 2018 & 2018 & 2017 & 2018 & 2019 & 2019 & 2020 & 2021\\
\hline
Sup. & \XSolidBrush & \XSolidBrush & \XSolidBrush & \XSolidBrush & \CheckmarkBold & \CheckmarkBold & \CheckmarkBold & \CheckmarkBold & \CheckmarkBold & \CheckmarkBold & \CheckmarkBold & \CheckmarkBold \\
\hline
Overall & 38.9 & 46.6 & 64.2 & \textbf{68.2} & 51.2 & 53.8 & 55.2 & 64.6 & 66.1 & 79.4 & 73.2 & \textbf{81.5} \\
\( \mathcal{J} \)\ Seen & 43.1 & 50.6 & 63.9 & \textbf{67.0} & 60.0 & 59.5 & 60.1 & 71.0 & 67.8 & 79.7 & 72.6 & \textbf{82.1} \\
\( \mathcal{F} \)\ Seen & 38.6 & 46.6 & 64.9 & \textbf{68.4} & 60.1 & - & 62.7 & 70.0 & - & 84.2 & 75.6 & \textbf{85.7} \\
\( \mathcal{J} \)\ Unseen & 36.6 & 43.8 & 60.3 & \textbf{64.5} & 40.6 & 45.2 & 46.6 & 55.5 & 60.8 & 72.8 & 68.9 & \textbf{75.7} \\
\( \mathcal{F} \)\ Unseen & 37.4 & 45.6 & 67.7 & \textbf{73.2} & 44.0 & - & 51.4 & 61.2 & - & 80.9 & 75.7 & \textbf{82.4} \\
Gen. Gap. & 3.9 & 3.9 & 0.4 & \textbf{-1.2} & 17.8 & 14.3 & 12.4 & 12.2 & 7.0 & 5.1 & \textbf{1.8} & 4.9 \\

\toprule[1.5pt]
\end{tabular}
\caption{Evaluation on YouTube-VOS 2018 validation set for ``seen" and ``unseen" categories (``unseen" object category does not appear in training). For \emph{overall}, \( \mathcal{J} \), and \( \mathcal{F} \), higher values are better and for Gen. (Generalization) Gap, lower values are better.}
\label{tabytbeval}
\end{table*}

\subsection{Evaluation Metrics}
We use Region Similarity \( \mathcal{J} \) and Countour Accuracy \( \mathcal{F} \) for evaluation. We also report the Generalization Gap as in \cite{MAST} to evaluate the generalization ability of MAMP on YouTube-VOS. Generalization Gap computes the model's performance gap between inference on seen and unseen categories. Its value is inversely proportional to the generalization ability of a model.

\subsection{Quantitative Results}

We compare MAMP with existing methods on DAVIS-2017 and YouTube-VOS 2018. Results are obtained using the official evaluation code and server. We do our best to compare the results as fairly as possible. For example, multi-stage training strategies, external datasets, data augmentations, and online adaptation are $not$ used in this work.

Table \ref{tabdaviseval} and Table \ref{tabytbeval} summarize the performance of the state-of-the-art methods and MAMP on DAVIS-2017 and YouTube-VOS 2018. MAMP significantly outperforms benchmark self-supervised methods by over 4.2\% on DAVIS-2017, 4\% on YouTube-VOS 2018, and by 4.85\% on the unseen categories of YouTube-VOS 2018. Moreover, MAMP is also comparable to some supervised methods. These results demonstrate the effectiveness of MAMP.

To evaluate the generalization ability of MAMP, we evaluate it on both ``seen" and ``unseen" categories of YouTube-VOS 2018. Objects in ``unseen" categories do not appear in the training set. Table \ref{tabytbeval} shows that MAMP performs well on ``unseen" categories and has the best generalization ability. Surprisingly, it performs better on ``unseen" categories than on ``seen" categories because of the better boundary segmentation performance on ``unseen" objects. These results indicate that MAMP can learn general feature representations that are not restricted by the specific object categories in the training set. The most comparable supervised method in generalization ability is GC \cite{GC} (1.8 vs -1.2). However, GC is trained with several external datasets with precise ground truth annotations.

In addition to YouTube-VOS 2018, we also evaluate MAMP on YouTube-VOS 2019 which has more videos and object instances. The results are included in supplementary material and show that MAMP achieves the best performance and generalization ability compared to other self-supervised methods on YouTube-VOS 2019 as well.

\begin{figure}[t] %  figure placement: here, top, bottom, or page
\centering
\includegraphics[width=\columnwidth]{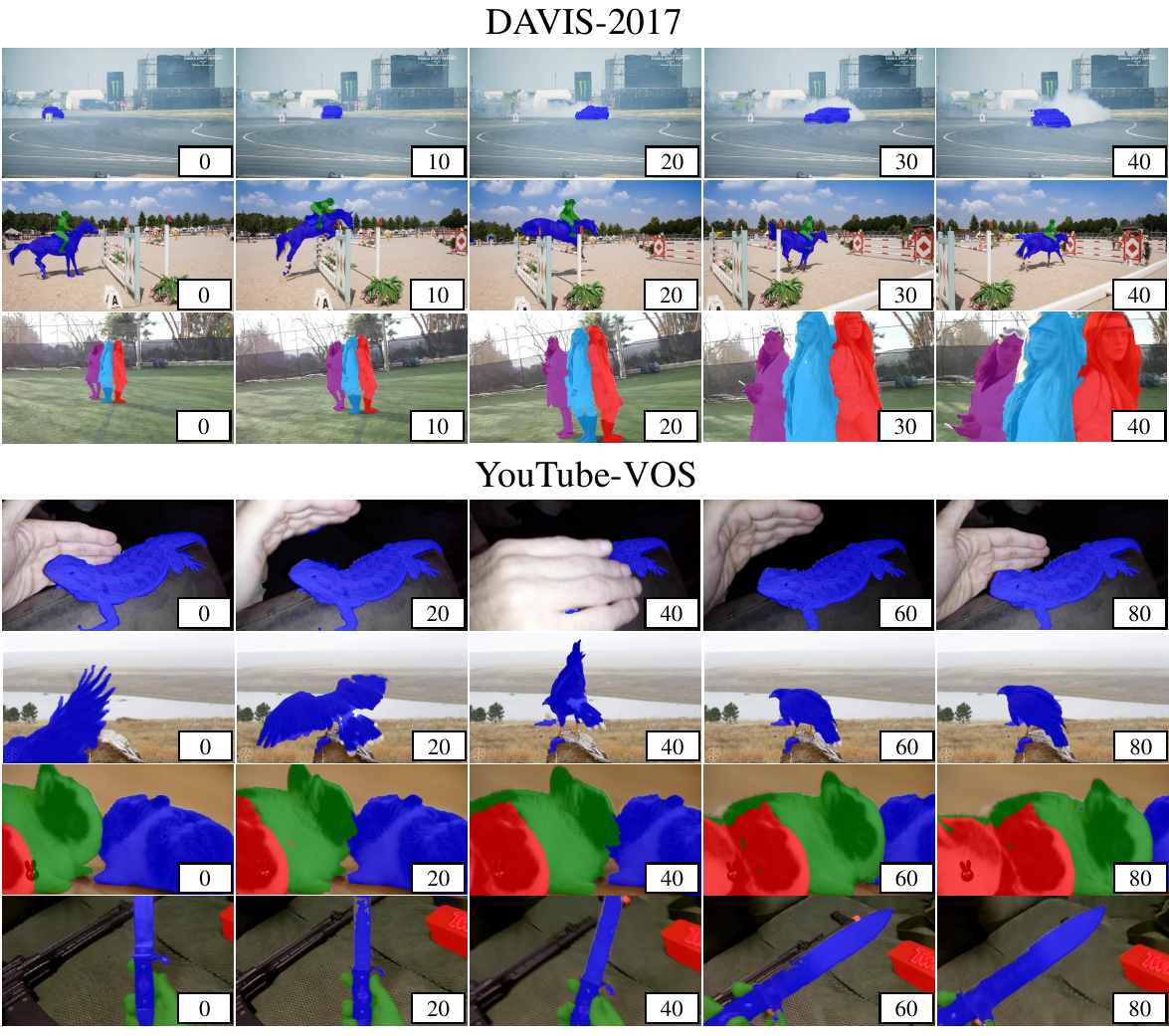}
\caption{Qualitative results of MAMP. The frames are sampled at fixed intervals and the first frame in each video is assigned index 0. MAMP performs well in challenging scenarios of occlusion/dis-occlusion, fast motion, large deformations, and scale variations.}
\label{fig:5}
\end{figure}

\begin{table}[t]
\small
\centering
\begin{tabular}{cccccc}
\toprule[1.5pt]
\multicolumn{1}{c}{M} & \multicolumn{1}{c}{A} & \multicolumn{3}{c}{DAVIS-2017} & \multicolumn{1}{c}{YouTube-VOS}\\
\cmidrule(lr){3-5}\cmidrule(lr){6-6} & & \( \mathcal{J} \)\ \& \( \mathcal{F} \)\ & \( \mathcal{J} \)\ & \( \mathcal{F} \)\ & \( \mathcal{J} \)\ \& \( \mathcal{F} \)\ \\
\hline
& & 65.8 & 63.6 & 68.0 & 61.8 \\
\hline
\checkmark & & 68.9 +3.1 & 66.9 +3.3 & 70.9 +2.9 & 68.2 +6.4 \\
& \checkmark & 67.4 +1.6 & 65.9 +2.3 & 68.9 +0.9 & 61.8 +0.0 \\
\checkmark & \checkmark & \textbf{69.7 +3.9} & \textbf{68.3 +4.7} & \textbf{71.2 +3.2} & \textbf{68.2 +6.4}\\
\toprule[1.5pt]
\end{tabular}
\caption{Ablation experiment for motion-aware spatio-temporal matching module (M) and size-aware image feature alignment module (A).}
\label{abmamp}
\end{table}

\subsection{Qualitative Results}

Figure \ref{fig:5} shows qualitative results of MAMP under various challenging scenarios, e.g., occlusion, fast motion, large deformations, and scale variations. MAMP is able to handle these challenging scenarios effectively.

\subsection{Ablation Studies}

\noindent\textbf{Motion-aware Spatio-temporal Matching.}
In Table \ref{abmamp}, we replaced the vanilla local spatio-temporal matching module of \cite{MAST} with the proposed motion-aware spatio-temporal matching module and the performance increased by 3.1\% on DAVIS-2017 and 6.4\% on YouTube-VOS 2018. Vanilla local spatio-temporal matching module retrieves the corresponding ROIs according to the pre-defined radius and ROI localization method which is not accurate enough as the most correlated ROIs are prone to be outside the search radius under fast motion and long-term matching scenarios. Without these most correlated ROIs, the label of one location in the query frame will be determined by the labels of several uncorrelated or weakly-correlated locations in the reference frames. Therefore, the segmentation results could be further improved if we can retrieve the most correlated ROIs for each location in the query frame.

Our motion-aware spatio-temporal matching module leverages the motion cues to register the reference frames in the memory bank to the query frame before computing the local spatio-temporal correspondences and filters out weakly correlated noise before mask propagation. Therefore, the above issues are alleviated even if the reference frames are temporally far from the query frame. As shown in Table \ref{abmamp}, the motion-aware spatio-temporal matching module brings more performance gains in YouTube-VOS 2018, as this dataset has longer video clips and lower frame rates compared to DAVIS-2017. To further demonstrate the effectiveness of our motion-aware spatio-temporal matching module, we choose different $k$ values to select the TopK correlated locations in the ROIs for mask propagation, the results show that MAMP still achieves the best performance compared to previous benchmarks even if only one of the 3125 locations in the ROIs is used for mask propagation (details in Supplementary Material).
\\

\noindent\textbf{Size-aware Image Feature Alignment.}
We removed the image feature alignment module of \cite{MAST} to replace it with our size-aware image feature alignment module. The performance increased by 1.6\% on DAVIS-2017 and remained unchanged on YouTube-VOS 2018 (see Table \ref{abmamp}). This is because our size-aware image feature alignment module fixes the misalignment at the upsampling stage caused by the improper input size. After computing the statics, we found that the number of videos that have an improper input size is 96.7\% for DAVIS-2017 validation set and only 1.9\% for the YouTube-VOS 2018 validation set.
\\

\noindent\textbf{Long-term Memory and Short-term Memory.} In Table \ref{abmem}, we compare results for different memory settings. We can see that all memory settings have reasonable performance. Long-term memory provides accurate ground truth information for query frames, while short-term memory offers up-to-date information from past neighboring frames. The results show that MAMP with short-term memory performs better than using long-term memory. This is because the appearance and scale of objects usually change significantly over time and long-term memory alone is unable to adapt to these changes. Furthermore, it can be seen that MAMP using both memory types has the best performance as both memories are complementary.

\section{Comparison with MAST}
Our nearest competitor is MAST \cite{MAST} which leverages local spatio-temporal matching module with ROI localization for long-term mask propagation. The framework of MAMP is inspired from MAST, however, MAMP is different in various aspects: (1) The local spatio-temporal matching in MAST is sub-optimal for handling fast motions and long-term matching scenarios whereas the proposed motion-aware spatio-temporal matching in MAMP can better handle (see Table \ref{abmamp}) such situations by exploiting motion cues and noise filters. (2) The local spatio-temporal matching module of MAST has a larger memory footprint compared to the proposed motion-aware spatio-temporal matching in MAMP. One 3090 GPU supports only 5 reference frames in the memory bank for MAST, but 18 reference frames for MAMP. (3) The image feature alignment module of MAST does not fix the misalignment at the upsampling stage, whereas the size-aware image feature alignment module of MAMP alleviates this problem as shown in Fig.~\ref{fig:4}(b) and Table \ref{abmamp}. (4) MAST leverages multi-stage training strategies for training, i.e., finetunes the model using multiple reference frames. However, MAMP is trained end-to-end with pairwise frames once only. (5) MAST does not converge well using new releases of PyTorch and only obtains about 50 mean \( \mathcal{J} \)\&\( \mathcal{F} \) on the validation set of DAVIS-2017 when training with PyTorch 1.9. MAMP solves this issue by normalizing the spatio-temporal affinity matrix with the channel number and achieves 69.7 mean \( \mathcal{J} \)\&\( \mathcal{F} \) on the validation set of DAVIS-2017. Finally, qualitative comparisons of MAST and MAMP further demonstrate the superiority of MAMP (see videos in Supplementary Material).

\begin{table}[t]
\small
\centering
\begin{tabular}{lcccc}
\toprule[1.5pt]
\multicolumn{1}{c}{Memory} & \multicolumn{3}{c}{DAVIS-2017} & \multicolumn{1}{c}{YouTube-VOS}\\
\cmidrule(lr){2-4}\cmidrule(lr){5-5} & \( \mathcal{J} \)\ \& \( \mathcal{F} \)\ & \( \mathcal{J} \)\ & \( \mathcal{F} \)\ & \( \mathcal{J} \)\ \& \( \mathcal{F} \)\ \\
\hline
Long \& Short & \textbf{69.7} & \textbf{68.3} & \textbf{71.2} & \textbf{68.2} \\
\hline
Long & 52.3 & 49.7 & 54.9 & 58.7 \\
Short & 64.1 & 62.5 & 65.7 & 66.4 \\
\toprule[1.5pt]
\end{tabular}
\caption{Ablation study for long and short-term memory.}
\label{abmem}
\end{table}

\section{Conclusion}
In this paper, we proposed MAMP that enables general feature representation and motion-guided mask propagation. MAMP model can be trained without the need for annotations, and outperforms existing self-supervised methods by a large margin. Moreover, MAMP demonstrates the best generalization ability compared to previous methods. We believe that MAMP has the potential to propagate spatio-temporal features and masks in practical video segmentation tasks. In the future, we will develop more effective pretext tasks and adaptive memory selection methods to further improve the performance of MAMP.

\section{Acknowledgements}
\vspace{-1mm}
This research was supported by the ARC Industrial Transformation Research Hub IH180100002.

% Use \bibliography{yourbibfile} instead or the References section will not appear in your paper
\bibliography{aaai22}

\newpage
\newpage
\appendix
\section{Backbone Architecture}
\vspace{2mm}
We modify ResNet-18 and use it as the encoder to extract image features with a spatial resolution of 1\verb|/|4 of the input images, the specific architecture is shown in Table \ref{resnet}.

\begin{table}[h]
\centering
\begin{tabular}{ccc}
\toprule[1.5pt]
Layer Name & Output Size & Configuration \\
\hline
Conv1 & $H/2 \times W/2$ & 7 $\times$ 7, 64, stride 2\\
Conv2 & $H/2 \times W/2$ & $
\begin{bmatrix} 3 \times 3, 64 \\ 3 \times 3, 64 \end{bmatrix} \times$ 2 \\
Conv3 & $H/4 \times W/4$ & $
\begin{bmatrix} 3 \times 3, 128 \\ 3 \times 3, 128 \end{bmatrix} \times$ 2 \\
Conv4 & $H/4 \times W/4$ & $
\begin{bmatrix} 3 \times 3, 256 \\ 3 \times 3, 256 \end{bmatrix} \times$ 2 \\
Conv5 & $H/4 \times W/4$ & $
\begin{bmatrix} 3 \times 3, 256 \\ 3 \times 3, 256 \end{bmatrix} \times$ 2 \\
\toprule[1.5pt]
\end{tabular}
\caption{Architecture of the modified ResNet-18.}
\label{resnet}
\end{table}

\section{TopK Correlated Locations for Mask Propagation}
\vspace{2mm}

If we retrieve ROIs with a radius of 12 on 5 reference frames, the corresponding ROIs for one location in the query frame will include 3125 locations. However, noisy matches to 3125 locations may adversely affect the model's performance. Hence, the proposed motion-aware spatio-temporal matching filters out redundant and noisy ROIs by selecting TopK correlated locations only for mask propagation. As shown in Table \ref{abtopk}, leveraging the top 36 or top 9 correlated locations in the ROIs for mask propagation improves the performance of MAMP compared to using all 3125 locations. Using the top 36 correlated locations obtains the best performance. Moreover, compared to other {state-of-the-art self-supervised} methods, MAMP still maintains the best performance even if only one of the 3125 locations in the ROIs is used for mask propagation. These results further demonstrate the effectiveness of the proposed motion-aware spatio-temporal matching module.

\begin{table}[h]
\small
\centering
\begin{tabular}{lcccc}
\toprule[1.5pt]
\multicolumn{1}{c}{TopK} & \multicolumn{3}{c}{DAVIS-2017} & \multicolumn{1}{c}{YouTube-VOS}\\
\cmidrule(lr){2-4}\cmidrule(lr){5-5} & \( \mathcal{J} \)\ \& \( \mathcal{F} \)\ & \( \mathcal{J} \)\ & \( \mathcal{F} \)\ & \( \mathcal{J} \)\ \& \( \mathcal{F} \)\ \\
\hline
ALL  & 69.0 & 67.6 & 70.3 & 67.9 \\
\hline
Top 1  & 66.5 -2.5 & 64.4 -3.2 & 68.6 -1.7 & 66.3 -1.6 \\
Top 9  & 69.4 +0.4 & 67.7 +0.1 & \textbf{71.2 +0.9} & \textbf{68.4 +0.5} \\
Top 36  & \textbf{69.7 +0.7} & \textbf{68.3 +0.7} & \textbf{71.2 +0.9} & 68.2 +0.3 \\
\toprule[1.5pt]
\end{tabular}
\caption{Ablation of TopK correlated locations in the ROIs for mask propagation.} \label{abtopk}
\end{table}

\section{Evaluation on YouTube-VOS 2019}
\vspace{2mm}

We also evaluate the proposed MAMP on YouTube-VOS 2019. YouTube-VOS 2019 includes more videos and object instances compared to YouTube-VOS 2018. As shown in Table \ref{ytb19}, MAMP still significantly outperforms other self-supervised methods and has the best generalization ability.

\begin{table}[h]
\small
\centering
\renewcommand\tabcolsep{4.0pt}
\begin{tabular}{lccccccc}
\toprule[1.5pt]
\multicolumn{1}{c}{Method} & \multicolumn{1}{c}{Sup.} & \multicolumn{1}{c}{Overall} & \multicolumn{2}{c}{Seen} & \multicolumn{2}{c}{Unseen} & \multicolumn{1}{c}{Gen. Gap}\\
\cmidrule(lr){4-5}\cmidrule(lr){6-7} & & &  \( \mathcal{J} \)\ & \( \mathcal{F} \)\ & \( \mathcal{J} \)\ & \( \mathcal{F} \)\ & \\
\hline
Vid. Color. & \XSolidBrush &  39.0 & 43.3 & 38.2 & 36.6 & 37.5 & 3.7 \\
CorrFlow & \XSolidBrush & 47.0 & 51.2 & 46.6 & 44.5 & 45.9 & 3.7 \\
MAST & \XSolidBrush & 64.9 & 64.3 & 65.3 & 61.5 & 68.4 & 0.15 \\
\textbf{Ours} & \XSolidBrush & \textbf{68.2} & \textbf{66.3} & \textbf{67.5} & \textbf{65.4} & \textbf{73.7} & \textbf{-2.6} \\
\toprule[1.5pt]
\end{tabular}
\caption{Evaluation on YouTube-VOS 2019 validation set for ``seen" and ``unseen" categories.}
\label{ytb19}
\end{table}

\end{document}